\def\year{2022}\relax
\documentclass[letterpaper]{article} 
\usepackage{aaai22}  
\usepackage{times}  
\usepackage{helvet}  
\usepackage{courier}  
\usepackage[hyphens]{url}  
\usepackage{graphicx} 
\usepackage{natbib}  
\usepackage{caption} 
\DeclareCaptionStyle{ruled}{labelfont=normalfont,labelsep=colon,strut=off} 
\usepackage{algorithm}
\usepackage{algorithmic}

\usepackage{newfloat}
\usepackage{listings}
\usepackage[switch]{lineno}
\usepackage{amsmath}
\usepackage{amsthm}
\usepackage{psfrag}
\usepackage{graphicx}
\usepackage{amssymb}
\usepackage{subfigure}
\usepackage{algorithmic}
\usepackage{algorithm}
\usepackage{cases}
\usepackage{ulem}
\usepackage{amsmath,amssymb}

\usepackage{color,url}
\usepackage{threeparttable}
\usepackage{booktabs}
\usepackage{slashbox}

\usepackage{textcomp}
\usepackage{epstopdf}
\usepackage{threeparttable}
\usepackage{booktabs}
\usepackage{bm}
\usepackage{appendix}

\usepackage[Symbol]{upgreek}
\usepackage{caption,setspace}
\usepackage{amsmath}
\newtheorem{theorem}{Theorem}

\newtheorem{definition}{Definition}
\newtheorem{assumption}{Assumption}
\newtheorem{condition}{Condition}

\floatstyle{ruled}
\newfloat{listing}{tb}{lst}{}
\floatname{listing}{Listing}
\title{Zero Stability Well Predicts Performance of Convolutional Neural Networks}
\author {
    Liangming Chen\textsuperscript{\rm 1, 2, 3}, 
    Long Jin\textsuperscript{\rm 1, 3}\thanks{Corresponding Author},
    Mingsheng Shang\textsuperscript{\rm 1}
}
\affiliations {
    \textsuperscript{\rm 1} Chongqing Key Laboratory of Big Data and Intelligent Computing, Chongqing Institute of Green and Intelligent Technology, Chinese Academy of Sciences\\
    \textsuperscript{\rm 2} Chongqing School, University of Chinese Academy of Sciences\\
    \textsuperscript{\rm 3} School of Information Science and Engineering, Lanzhou University\\
    \{lmchen, jinlongsysu\}@foxmail.com, msshang@cigit.ac.cn
}
\usepackage{bibentry}

\begin{document}

\maketitle

\begin{abstract}
    The question of what kind of convolutional neural network (CNN) structure performs well is fascinating. In this work, we move toward the answer with one more step by connecting zero stability and model performance. Specifically, we found that if a discrete solver of an ordinary differential equation is zero stable, the CNN corresponding to that solver performs well. We first give the interpretation of zero stability in the context of deep learning and then investigate the performance of existing first- and second-order CNNs under different zero-stable circumstances. Based on the preliminary observation, we provide a higher-order discretization to construct CNNs and then propose a zero-stable network (ZeroSNet). To guarantee zero stability of the ZeroSNet, we first deduce a structure that meets consistency conditions and then give a zero stable region of a training-free parameter. By analyzing the roots of a characteristic equation, we theoretically obtain the optimal coefficients of feature maps. Empirically, we present our results from three aspects: We provide extensive empirical evidence of different depth on different datasets to show that the moduli of the characteristic equation's roots are the keys for the performance of CNNs that require historical features; Our experiments show that ZeroSNet outperforms existing CNNs which is based on high-order discretization; ZeroSNets show better robustness against noises on the input. The source code is available at \url{https://github.com/LongJin-lab/ZeroSNet}. 
\end{abstract}

\section{Introduction}\label{intro}
\begin{figure}[th]
	\centering
	\subfigure[A-stability]{
		\label{figStab_a}
		\includegraphics[height=2.1in]{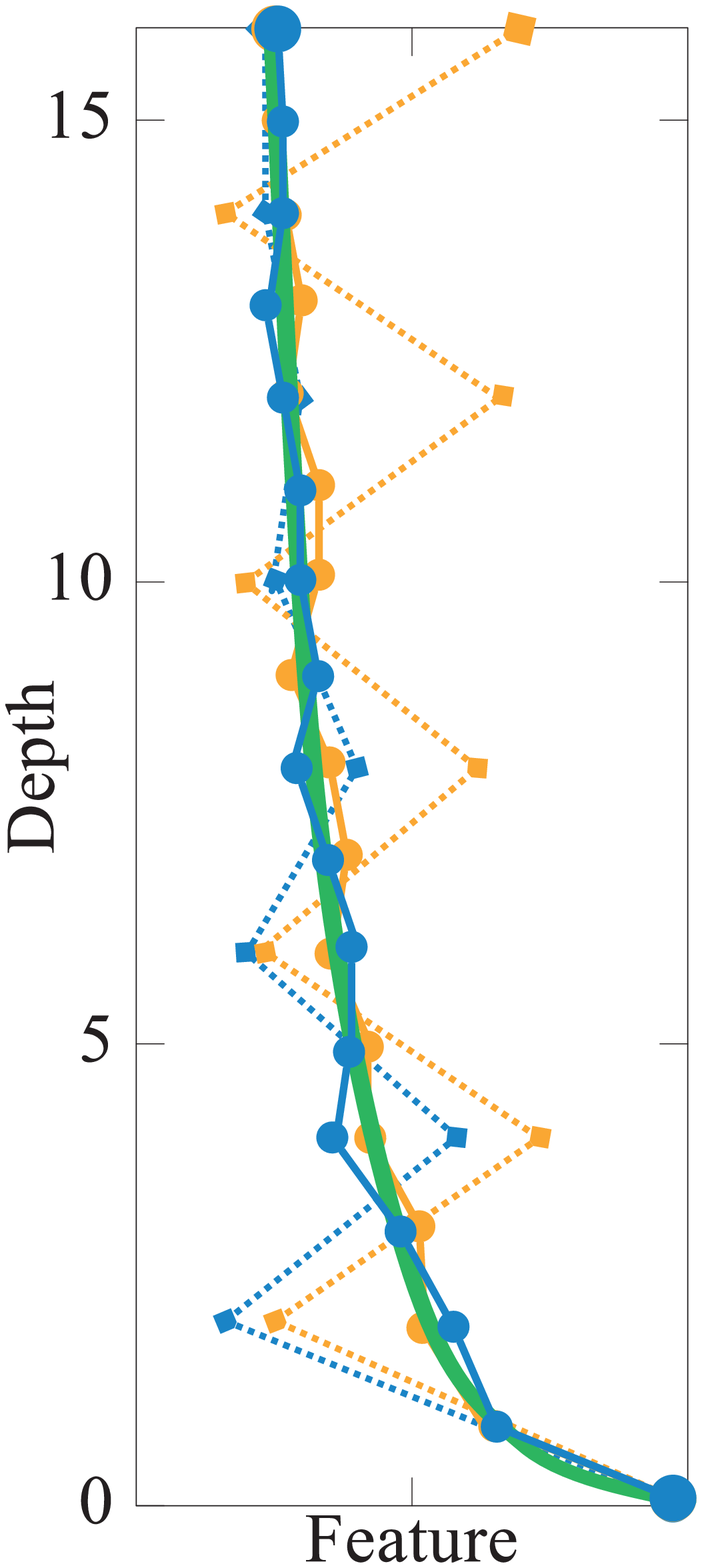}
	}
	\subfigure[BIBO stability]{
	\label{figStab_b}
	\includegraphics[height=2.1in]{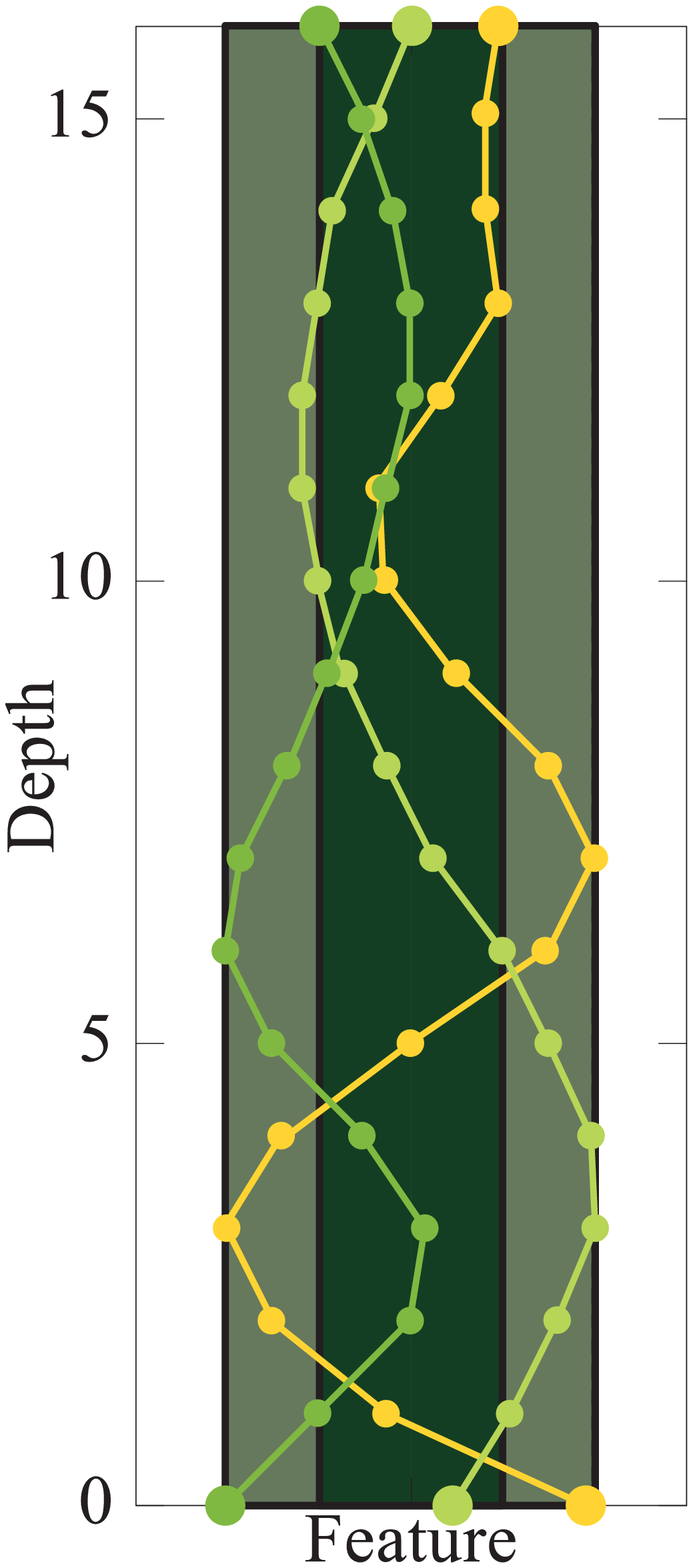}
}
\subfigure[Zero stability]{
	\label{figStab_c}
	\includegraphics[height=2.1in]{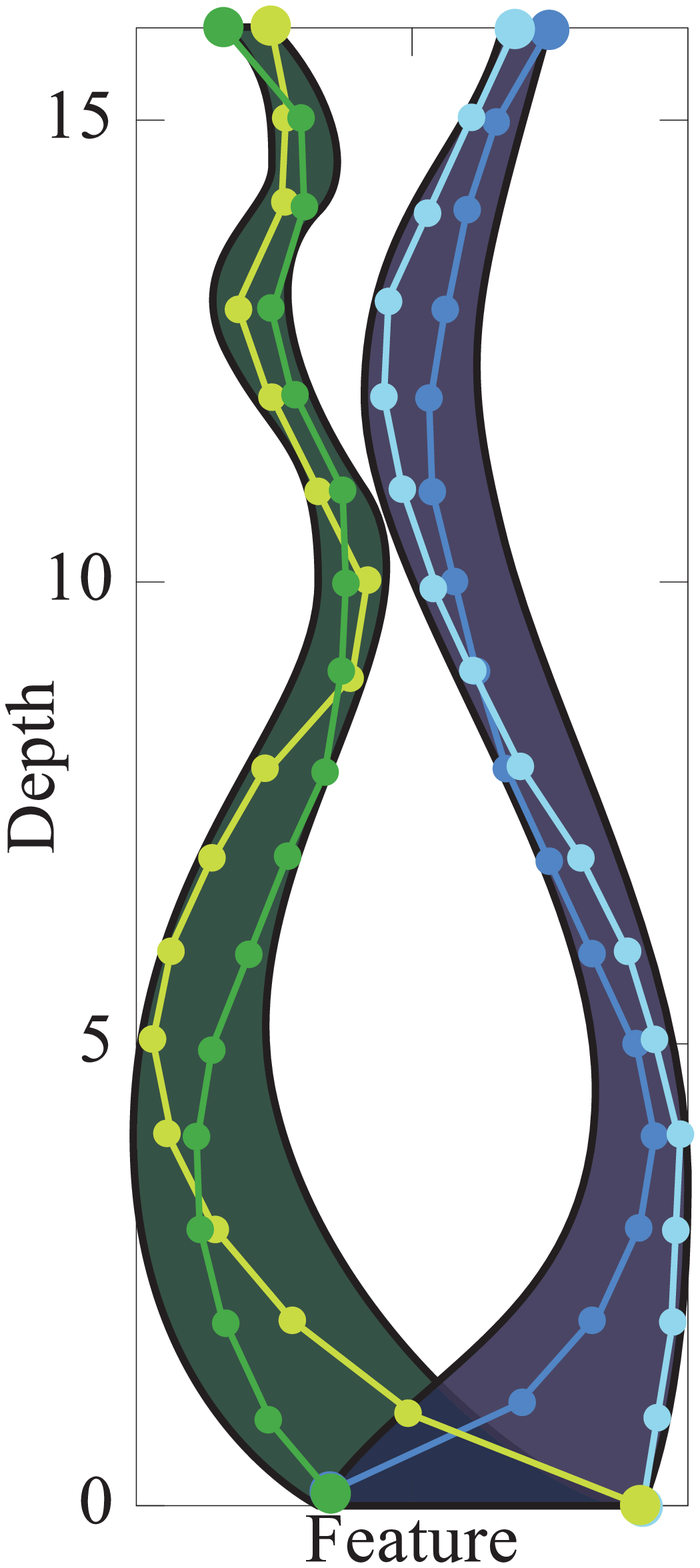}
}

	\caption{Illustrations of A-stability, BIBO stability, and zero stability. (a) Blue lines denote an A-stable method: Regardless of the step size, the method approaches the exact solution (the solid green curve). Orange lines represent a non-A-stable method, which can only approach the exact solution if the step size is small. Note that dotted lines have a large step size. (b) The light shade represents the bound of the input; The dark shade represents the bound of the output. (c) The shades represent possible ranges of the difference magnitude between features with different initial values. It means that similar inputs generate similar outputs. In this work, we focus on zero stability and connect it with robustness and generalization. }
	\label{figStab}
\end{figure}

The structure of a convolutional neural network (CNN) significantly affects its performance \cite{he2016identity, xie2019exploring, 9526915}. However, there is no clear clue about determining the importance of historical features and current activations (e.g., a sequence consisting of ReLU, convolutional layer, and batch normalization layer). A promising direction for structure determination is the ordinary-differential-equation-inspired design \cite{lu2018beyond, zhu2019convolutional}. We seek the answer from the perspective of zero stability which is a concept originating from numerical analysis.
\begin{figure*}[ht]	
	\centering
	\includegraphics[width=6.9in]{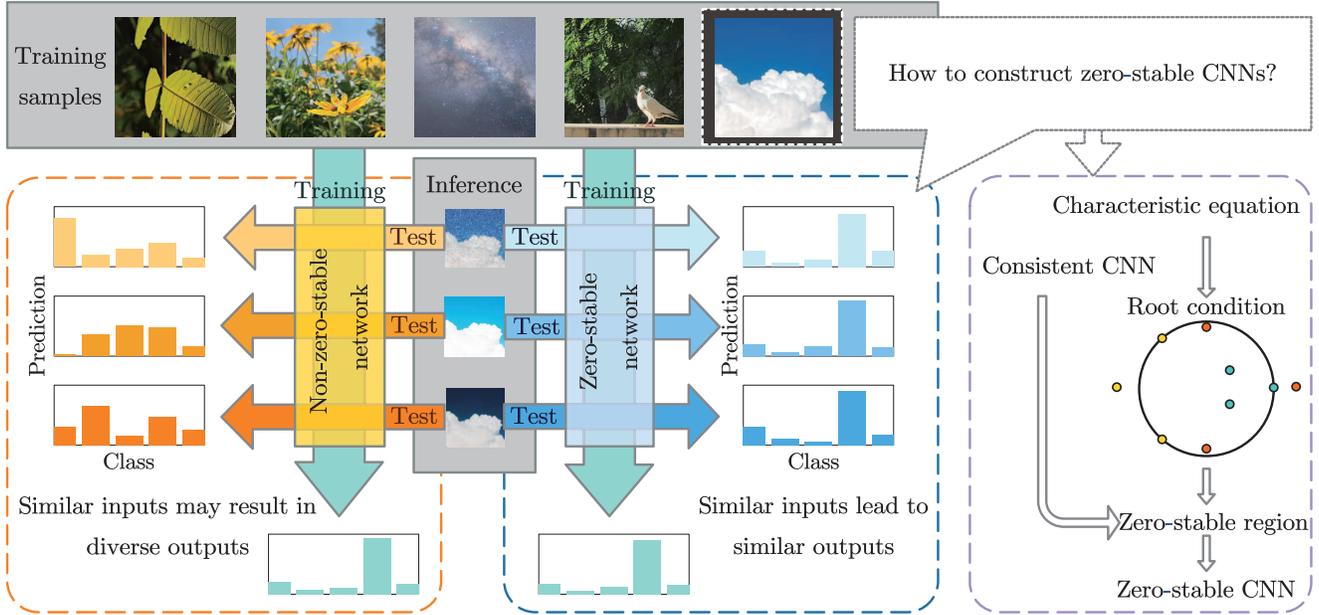}
	\caption{Left: Illustrations of the connection between zero stability and generalization/robustness. Right: Our approach to construct a zero-stable CNN, namely ZeroSNet. }
    \label{flowchart}
\end{figure*}

There are several types of stabilities in different fields. We provide Fig. \ref{figStab} to illustrate three of them: Absolute stability (A-stability), bounded input bounded output (BIBO) stability, and zero stability. In the following content, we discuss stability from a CNN structure design perspective. 
As shown in Fig. \ref{figStab_a}, absolute stability (A-stability) means that the numerical solution approaches the exact solution, regardless of the step size, as $t \rightarrow \infty$ \cite{atkinson2011numerical, 9205688}. For CNNs, A-stability means that for a network with a given depth, the feature map magnitude does not increase from the input to the output \cite{haber2019imexnet}. BIBO stability ensures that the output feature map is within a bound if the input is bounded \cite{zhang2020forward}, as in Fig. \ref{figStab_b}.
Although these stabilities benefit the learning ability and prevent the model from collapsing \cite{9684931, 9340571}, they give no clear clues about the CNNs' generalization and robustness. 


We show two meanings of zero stability in the context of numerical analysis and connect them with the CNNs' generalization and robustness, respectively. As shown in Fig. \ref{figStab_c}, the first meaning of zero stability is that for two similar initial values, the corresponding states at time $t$ are also similar \cite{gautschi1997numerical}. We borrow this idea to analyze the generalization of CNN: For a well-trained CNN, zero stability means that if the test sample is slightly different from the training one, the feature map changes slightly compared with that of the training sample, and then the network output also changes slightly. Since this CNN is well-trained as aforementioned, a correct prediction for the test sample should be given. Another meaning of zero stability is that if an initial value is perturbed, the fluctuation of the output is bounded \cite{atkinson2011numerical}, as in Fig. \ref{figStab_c}. We add the noise on the input feature as the perturbation on the initial value, thereby building a bridge between zero stability and the robustness of CNNs. 

To illustrate the insight of zero stability in the context of CNN, we give an example on the left side of Fig. \ref{flowchart}. Both zero-stable and non-zero-stable CNNs classify the cloud image correctly. When similar samples are fed into the two CNNs, the non-zero-stable CNN gives diverse predictions, while the zero-stable one gives similar predictions and thus succeeds in this task. To achieve zero stability, we propose a zero-stable network (ZeroSNet) with a general form to ensure consistency (which tightens the upper bound of zero stability) of the ZeroSNet. Based on the characteristic equation of the ZeroSNet, we apply the root condition and then obtain a zero-stable region for a flexible coefficient. The right side of Fig. \ref{flowchart} describes the process of constructing the ZeroSNet.



Our contributions in this paper are summarized as follows:
\begin{itemize}
	\item We are the first to find that zero stability well predicts the model performance (we provide preliminary observations and more convincing evidence). Based on the finding, we provide corresponding explanatory analyses. 
	\item We propose a CNN named ZeroSNet with theoretical proofs on its consistency and give a stability region of a training-free parameter. Besides, we deduce optimal coefficients for historical features and the current activations. 
	\item ZeroSNet with theoretically optimal coefficients achieves advanced performance and outperforms the existing high-order-discretization CNNs.
	\item Our experiments show that involved zero-stable CNNs are robust against noises that are injected on the input, while non-zero-stable ones reveal a dramatical degradation.
\end{itemize}

\section{Preliminaries}\label{Preliminaries}

We slightly extend the initial value problem \cite{atkinson2011numerical, chen2018neural} and get an initial values problem with more initial steps.
\begin{definition}[Initial values problem]
An initial values problem is defined as
    \begin{equation}\label{IVP}
        \begin{aligned}
        & \frac{\mathrm{d} \boldsymbol{y}(t) }{\mathrm{d} t}= \boldsymbol{f}(t,  \boldsymbol{y}(t)), \quad s \leq t \leq e, \\ & \boldsymbol{y}(s+qh)= \boldsymbol{y}_{s+q}, \quad q = 0, 1, \ldots, d,
    \end{aligned}
    \end{equation}
where $\boldsymbol{y}(t) \in \mathbb{R}^p$ is a $p$-dimensional feature vector; $s$ and $e$ are the start and the end of the time $t$, respectively; $h$ is the step size with $q$ denoting the $q$th step; $\boldsymbol{y}_{s+q-1}$ are given initial states. 
\end{definition}
In the context of the deep learning, $\boldsymbol{y}_{s}$ can be seen as the input of a neural network, and the training process determines the optimal $\boldsymbol{f}(t,  \boldsymbol{y}(t))$ \cite{lu2018beyond}. To discretize \eqref{IVP}, the definition of the $d$th-order discretization is given as follow.
\begin{definition}[$d$th-order discretization]
    A $d$th-order discretization for an initial value problem is defined as
    \begin{equation}\label{dOrder}
        \begin{aligned}
        \boldsymbol{y}(t_{n+1})= & \alpha_{0} \boldsymbol{y}_{n} + \alpha_1  \boldsymbol{y}(t_{n-1}) +\ldots + \alpha_{d-1} \boldsymbol{y}(t_{n-d+1}) \\
        & + h \beta \boldsymbol{f}(t_{n},\boldsymbol{y}(t_n) ), \\ &  n=0, 1,  \ldots, \quad d=1, 2, \ldots,
        \end{aligned}\\
        \end{equation}
        where $\alpha_0$, $\alpha_1$, $\ldots$, $\alpha_{d-1}$, and $\beta$ are given coefficients; $ t_{n}=s+n h$. 
    \end{definition}
    We can interpret $\alpha_0$, $\alpha_1$, $\ldots$, $\alpha_{d-1}$ and $\beta$ as the weight of each historical featuremap and the current activations, respectively. If $d=1$ and $\alpha_0=\beta=1$, equation \eqref{dOrder} reveals the Euler discretization, and it gives the pre-activation ResNet (PreResNet) \cite{he2016identity}.
    Moreover, equation \eqref{dOrder} can be regarded as a special case of the multilstep method. For $d=2$, if $\alpha_0=1-k_n$, $\alpha_1=k_n$, and $\beta=1$, one gets the linear multilstep (LM) architecture \cite{lu2018beyond}. In this work, we build higher-order-discretization-based CNNs, of which zero stability and consistency are guaranteed.
    \begin{assumption}[Lipschitz continuous sequence after normalization]\label{AssLip}
        Consider an $\boldsymbol{f}$ which consists of a sequence of layers (e.g., ReLU and convolutional layers) and a normalization layer in the end, and $\boldsymbol{f}$ is Lipschitz continuous. That is, for two arbitrary $\boldsymbol{y}, \hat{\boldsymbol{y}} \in \mathbb{R}^{p}$,
            \begin{equation}\label{EQLip}
                \left\|\boldsymbol{f}(t, \boldsymbol{y})-\boldsymbol{f}\left(t, \hat{\boldsymbol{y}}\right)\right\| \leq \ell \left\|\boldsymbol{y}-\hat{\boldsymbol{y}}\right\|, t \in [s, e],
            \end{equation}
            where $\ell$ is the Lipschitz constant; $\| \cdot \|$ denotes the 2-norm of a vector.
        \end{assumption}
        Usually, once the normalization layer (e.g., batch normalization, layer normalization) is involed as the last layer in $\boldsymbol{f}$, condition \eqref{EQLip} is meet for CNNs. This is because no matter how large the original feature values are, after a normalization layer, these values are forced to follow a controled distribution. An example of such a sequence in $\boldsymbol{f}$ is $\boldsymbol{f}'(t, \boldsymbol{y}) = \boldsymbol{n}_2 (ReLU( \boldsymbol{\theta}_2\star  \boldsymbol{n}_1(ReLU(\boldsymbol{\theta}_1 \star \boldsymbol{y}))))$, where $\boldsymbol{n}_1$ and $\boldsymbol{n}_2$ are both batch normalization layers; symbol $\star$ denotes the convolution operator. 
    \begin{definition}[Zero stablility \cite{gautschi1997numerical}]\label{DefZS}
        For two grid functions $\boldsymbol{y}$ and $\hat{\boldsymbol{y} }$ on $[a, b]$, a $d$th-order discretization is zero-stable if the following inequality holds for a sufficient-small step size $h$:
        \begin{equation}\label{EQZS}
            \begin{aligned}
             \|\boldsymbol{y}_{n}\!-\!\hat{\boldsymbol{y}}_{n}\|_{\infty} \leq\!\! c (\max _{ m \in [0, d-1]}\|\boldsymbol{y}_{m}\!\!-\!\!\hat{\boldsymbol{y}}_{m}\|\!+\!\| \boldsymbol{r} (\boldsymbol{y}_{n})\!-\!\boldsymbol{r} (\hat{\boldsymbol{y}}_{n}) \|_{\infty} ),
            \end{aligned}
        \end{equation}
        where $\boldsymbol{r} (\boldsymbol{y}_{n})\!\!:=\!\!{1}/{h} \sum_{i=0}^{d-1} \alpha_{i} \boldsymbol{y}_{n+i}\!-\!\beta \boldsymbol{f} \left(t_{n}, \boldsymbol{y}(t_{n})\right) $; $\boldsymbol{r} (\hat{\boldsymbol{y}}_{n})\!:=\!{1}/{h} \sum_{i=0}^{d-1} \alpha_{i} \hat{\boldsymbol{y}}_{n+i}\!-\!\beta \boldsymbol{f} \left(t_{n}, \hat{\boldsymbol{y}}(t_{n})\right)$; $c$ is a constant; $\| \cdot \|_{\infty}$ is the infinity norm.
    \end{definition}

    \begin{definition}[Consistency \cite{atkinson2011numerical}]\label{DefCons}
        For an exact solution $\boldsymbol{y}(t)$, a $d$th-order discretization is consistent if
        \begin{small}
        \begin{equation}\label{EQcons}
        \begin{aligned}
        &\max_{t_{n}\in[t_{d-1}, e]}  \left\| \boldsymbol{y}(t_{n+1})- \left( \sum_{i=0}^{d-1} \alpha_{i} \boldsymbol{y}_{n-i} + h \beta \boldsymbol{f}(t_{n},\boldsymbol{y}(t_n) ) \right) \right\|/h \\ &\rightarrow 0 \textrm{ as } h \rightarrow 0.
        \end{aligned}
        \end{equation}
        \end{small}
    \end{definition}
    \noindent If $\boldsymbol{y}$ and $\hat{\boldsymbol{y}}$ are two solutions with different initial values, $\boldsymbol{r} (\boldsymbol{y}_{n})$ and $\boldsymbol{r} (\hat{\boldsymbol{y}}_{n})$ are truncation errors exactly \cite{gautschi1997numerical}.
    If the $d$th-order discretization \eqref{dOrder} is consistent and the step size $h$ is sufficient-small , we have $\lim_{n \rightarrow \infty} \boldsymbol{r} (\boldsymbol{y}_{n})  \rightarrow \boldsymbol{0}$ and $\lim_{n \rightarrow \infty} \boldsymbol{r} (\hat{\boldsymbol{y}}_{n}) \rightarrow \boldsymbol{0}$ \cite{gautschi1997numerical}. Under the consistency condition \eqref{EQcons}, it follows that
    \begin{equation}\label{ZS_solution}
      \|\boldsymbol{y}_{n}\!-\!\hat{\boldsymbol{y}}_{n}\|_{\infty} \leq c\left(\max _{m \in [0, d-1]}\left\|\boldsymbol{y}_{m}-\hat{\boldsymbol{y}}_{m}\right\|\right).
    \end{equation}

\subsection{Criterion for Zero Stability}
Root condition, a well-known criterion for zero stability, is given here and is further as a practical tool to verify zero stability and predict the performance of CNNs later. 
\begin{condition}[Root Condition \cite{ascher1998computer}]
    \label{rootcond}
    The root condition means that the roots of a characteristic equation $r(\rho) = \rho^d - \sum_{i=0}^{d-1}\alpha_i \rho^{d-1-i}$ satisfy $|\rho_i| \leq 1$, and if $|\rho_i| = 1$ then $\rho_i$ is a simple root, where $|\cdot|$ denoting to take the modulus of a complex number.
\end{condition}

The empirical observations in the next section show that some existing CNNs can be interpreted as first- and second-order discretizations. After that, we construct a higher-order CNN to further verify the relationship between model performance and zero stability.

\section{Observations from Existing CNNs}\label{Observations}

Involving historical feature maps benefits the CNNs' representation ability \cite{huang2017densely}. Meanwhile, a visualization study suggests that historical features may smooth the loss landscape. However, the importance of each historical feature for the CNNs' performance remains unclear: We adjust the coefficients (weights) of historical feature maps and current activations in the provided preparatory experiments. 

\subsection{An Observation from PreResNet}
\begin{table}[bt]
  \centering
    \begin{tabular}{l|c|c|c}
    \hline       \hline
    Model & $\alpha$  & Z. S. & Test acc. (\%) \\
    \hline \hline
    PreResNet-32 & 2          & No        & 79.13$\pm$0.30 \\
    PreResNet-32 & 1.5        & No        & 87.07$\pm$0.14 \\
    \hline
    PreResNet-32 & 0.5        & Yes    & 92.52$\pm$0.42 \\
    PreResNet-32 & 0.7        & Yes     & 93.16$\pm$0.13 \\
    PreResNet-32 & 1          & Yes  & \boldmath{}\textbf{93.19}$\pm$0.17\unboldmath{} \\
    \hline \hline
    \end{tabular}%
    \caption{Test accuracies (mean $\pm$ standard deviation) obtained by setting different feature weight $\alpha$ on CIFAR-10 dataset. ``Z. S.'' and ``acc.'' denote zero stability and the accuracy, respectively. Once the zero stability region is exceeded, the performance shows a clear degradation. }
  \label{1OrderAcc}%
\end{table}%

As discussed earlier, PreResNet can be deemed as an Euler discretization. Extending the Euler discretization slightly by involving a flexible coefficient $\alpha$ for the current feature $\boldsymbol{y}_{n}$ gives
\begin{equation}
    \boldsymbol{y}_{n+1} = \alpha \boldsymbol{y}_{n} + h \boldsymbol{f}(t_n,  \boldsymbol{y}_n).
\end{equation}
Let us see what happens if we change the value of $\alpha$ from Table \ref{1OrderAcc}. According to \cite{gautschi1997numerical}, we can check the stability quickly.
As shown in Table \ref{1OrderAcc}, there is a significant gap in the test accuracy between zero-stable and non-zero-stable models. Besides, the original PreResNet ($\alpha=1$) outperforms other models with the same structures but different coefficients $\alpha$.

We are still not sure whether the phenomenon is caused by the forward propagation or the backward propagation (from the backward propagation perspective, PreResNet may also benefit from the residual connection when applying the chain rule, as discussed in Section 3 of \cite{he2016identity}). We consider a second-order situation in the following subsection. 

\subsection{An Observation from LM-Architecture}

The LM-architecture in \cite{lu2018beyond} can be seen as a second-order discretization. We make a  modification on $\beta$ with a sharing $k$ for all layers to ensure consistency and then obtain
\begin{equation}
    \boldsymbol{y}_{n+1} = (1-k) \boldsymbol{y}_{n} + k \boldsymbol{y}_{n-1} + (2k+1) h \boldsymbol{f}(t_n, \boldsymbol{y}).
    \label{eq:LM}
\end{equation}
The characteristic equation of equation \eqref{eq:LM} is
\begin{equation}
    \label{root2}
    r(\rho) = \rho^2 + (k-1)\rho - k.
\end{equation}
We set several $k$ and check zero stability by applying the root condition \cite{atkinson2011numerical} for equation \eqref{root2}, and then obtain Table \ref{tab:2ndAcc}. The detailed experiment settings are described in Experiment section.
\begin{table}[tb]
    \centering
      \begin{tabular}{l|c|c|c}
      \hline \hline
      Model & $k$ & Z. S. & Test acc. (\%)\\
      \hline \hline
      LM-ResNet-44 & -1.5    & No     & 81.43$\pm$0.19 \\
      LM-ResNet-44 & 1.5     & No     & 89.46$\pm$0.30 \\
      \hline
      LM-ResNet-44 &  -0.5  & Yes  & 92.95$\pm$0.24 \\
      LM-ResNet-44 & 0.5   & Yes  & \boldmath{}\textbf{93.69}$\pm$0.21\unboldmath{} \\
      \hline \hline
      \end{tabular}%
    \caption{Test accuracies (mean $\pm$ standard deviation) obtained by setting different $k$ on CIFAR-10 dataset. ``Z. S.'' and ``acc.'' denote zero stability and the accuracy, respectively. Similar with the first-order discretization, the second-order discretization's performance can also be predicted by zero stability. }
    \label{tab:2ndAcc}%
  \end{table}%

\subsection{Zero Stability for CNN}\label{explain}

For CNNs, the meaning of equation \eqref{ZS_solution} is as follows. First, the backpropagation determines $\boldsymbol{f}$. When the network training process is done, we obtain $\boldsymbol{f}$, which fits the training data. We use $\boldsymbol{y}_0$ and $\hat{\boldsymbol{y}}_0$ to represent the inputs from the training set and the test set, respectively. From equation \eqref{ZS_solution}, if the inputs $\boldsymbol{y}_0$ and $\hat{\boldsymbol{y}}_0$ are similar, the predictions $\boldsymbol{y}_{n}$ and $\hat{\boldsymbol{y}}_{n}$ are similar, too. Assume that for a well-trained network, feeding $\boldsymbol{y}_0$ to the equation \eqref{dOrder} gives the correct prediction $\boldsymbol{y}_{n}$. Then, we could conclude that for a test sample, which is similar to one of the training samples, the prediction result is close to the correct answer. It means that the zero-stable neural networks are robust against perturbations, and its generalization is well for value-based differences.

To further investigate whether zero stability well predicts the model performance, we deduce a consistent and zero-stable model named ZeroSNet in the next section and practically compare it with its non-zero-stable counterparts in the Experiment section.



\section{Zero-Stable Network (ZeroSNet)}\label{SEC_ZeroSNet}

In this section, we deduce a CNN named Zero-Stable Network (ZeroSNet), which is automatically consistent, and we give the range of a flexible and training-free parameter to ensure zero stability.

\begin{figure}[bt]	
	\centering
	\includegraphics[width=0.48 \textwidth]{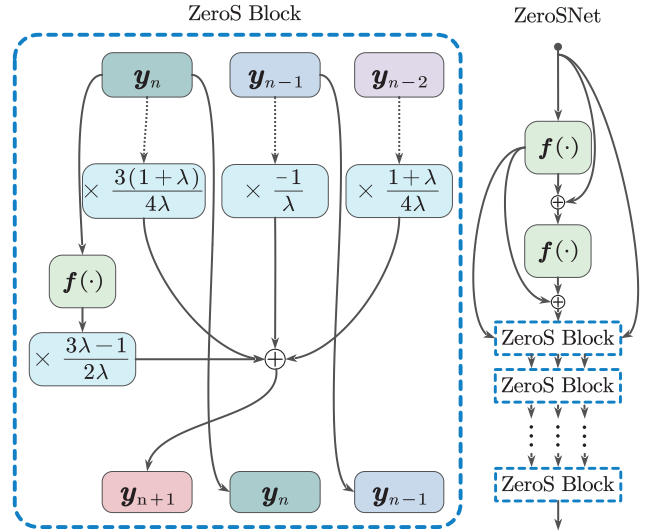}
	\caption{An overall structure of the ZeroSNet. Left: Two types of blocks with given coefficients that guarantee consistency and zero stability. The coefficients are given by equation \eqref{EQyp2} and are training-free. Note that each block takes three inputs and gives three outputs. The dotted arrow means performing downsampling if dimensions are mismatched. Left: A zero-stable block (ZeroS block). Right: Following the two basic start blocks, we stack ZeroS blocks to build the deep structure. $\boldsymbol{f}(\cdot)$ in the basic ZeroS block consists of two BN-ReLU-convolution triplets, and $\boldsymbol{f}(\cdot)$ in the ZeroS bottleneck block consists of three BN-ReLU-convolution triplets. }
    \label{FigZNet}
\end{figure}

\subsection{Description of ZeroSNet}
\begin{figure*}[tb]
	\centering
		\includegraphics[width=0.8 \textwidth]{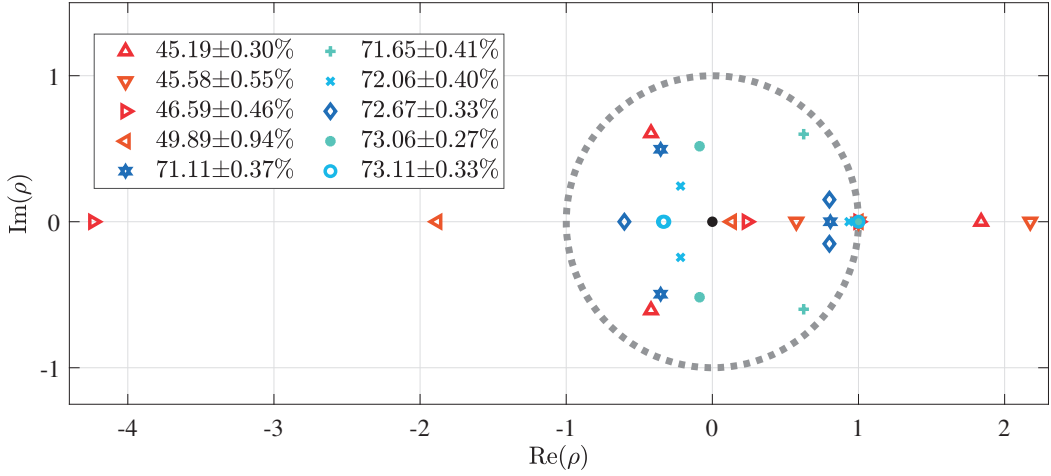}
	\caption{Test accuracies (mean $\pm$ standard deviation) of third-order-discretization-based models with different roots. The models all contains 56 layers and are evaluated on CIFAR-100 dataset. It is clear that models which satisfy the root condition (Condition \ref{rootcond}) outperform the ones with large roots. }
	\label{RootsAndAccC100L56}
\end{figure*}
In \cite{8423727}, a numerical method named general square-pattern discretization is presented. With the aid of this numerical method, we construct the ZeroSNet. 
Due to the space limitation, a mathematical derivation of the ZeroSNet is given in the Appendix. Directly, we give the formal description of the ZeroSNet here: 
\begin{equation}\label{EQyp2}
    \begin{aligned}
    \boldsymbol{y}_{n+1} = &\frac{3(1 + \lambda)}{4 \lambda}\boldsymbol{y}_n - \frac{1}{\lambda} \boldsymbol{y}_{n-1} + \frac{1 + \lambda}{4 \lambda} \boldsymbol{y}_{n-2} \\ & + \frac{3\lambda - 1}{2\lambda}h\boldsymbol{f}(t_n, \boldsymbol{y}_n),
    \end{aligned}
\end{equation}
where $\lambda \neq 0$ is a scalar.

To illustrate the structure of ZeroSNet, we provide Fig. \ref{FigZNet}. We borrow ideas from \cite{he2016identity} and \cite{zhang2017polynet}: Building a deep network by stacking blocks iteratively. 

Note that each block takes three inputs and gives three outputs. In Fig. \ref{FigZNet}, the dotted arrow means to perform a downsampling if dimensions of $\boldsymbol{y}_n, \boldsymbol{y}_{n-1}$, and $\boldsymbol{y}_{n-2}$ are mismatched. Three arrows pointing to the ``ZeroS Block/Bottlenecks'' module are respectively $\boldsymbol{y}_n, \boldsymbol{y}_{n-1}$, and $\boldsymbol{y}_{n-2}$; Arrows starting from the ``ZeroS Block/Bottlenecks'' module are respectively $\boldsymbol{y}_{n+1}, \boldsymbol{y}_{n}$, and $\boldsymbol{y}_{n-1}$. As will be discussed, Blocks in Fig. \ref{FigZNet} ensure consistency according to the following Theorem \ref{ThCons}. Moreover, if we follow Theorem \ref{THEZS} to set $\lambda$, the neural network is zero-stable. Compared to the PreResNet, the additional parameters of ZeroSNet are only for downsampling. We provide a comparison of the number of parameters in the Experiment section. By setting $\lambda_n$ as a trainable parameter for the $n$th block ($n \geq 2$), we obtain a trainable version of the ZeroSNet, and we call it ZeroSNet-Tra. Theorems \ref{ThCons} and \ref{THEZS}, as well as other properties of the ZeroSNet, will be introduced in the next section.

\subsection{Properties of ZeroSNet}
Compared to the PreResNet and the LM-ResNet, ZeroSNet involves more historical information and smoothly transmits the low-level features. However, \cite{gautschi1997numerical} suggests that high-order discretizations are easily to be non-zero-stable. 
 As shown in equation \eqref{ZS_solution}, consistency removes the $\| \boldsymbol{r}(\boldsymbol{y}_n)-\boldsymbol{r}(\hat{\boldsymbol{y}}_n) \|$ term in equation \eqref{EQZS} and thus gives a tighter upper bound of zero stability. Based on equation \eqref{ZS_solution}, zero stability has an ability to predict the model performance.
In this part, we give Theorem \ref{ThCons} to ensure consistency of the ZeroSNet and then provide a zero stability region of parameter $\lambda$.
\begin{theorem}[ZeroSNet \eqref{EQyp2} is consistent] \label{ThCons}
    Suppose that $\boldsymbol{y}(t)$ is continuously differentiable, ZeroSNet \eqref{EQyp2} meets the consistency condition. 
\end{theorem}

Proofs are deferred to the Appendix. 
By using the root condition \cite{atkinson2011numerical}, we investigate zero stability of ZeroSNet \eqref{EQyp2}.
\begin{theorem}[Zero stability region of the ZeroSNet \eqref{EQyp2}]
    \label{THEZS}
    For a continuously differentiable function $\boldsymbol{y}(t)$, if $\lambda \in (- \infty, -1) \cup (1/3, +\infty)$, the ZeroSNet \eqref{EQyp2} is zero-stable.
\end{theorem}
Proofs are deferred to the Appendix.
Based on Theorem \ref{THEZS}, we show optimal coefficients of historical features and the current activations $\boldsymbol{f}$.
\begin{theorem}[Optimal coefficients of the ZeroSNet \eqref{EQyp2}]
    \label{THEOW}
    From the perspective of zero stability, optimal coefficients of $ \boldsymbol{y}_n, \boldsymbol{y}_{n-1}, \boldsymbol{y}_{n-2} $, and $\boldsymbol{f}(t_n, \boldsymbol{y}_n)$ in the ZeroSNet \eqref{EQyp2} are $1/3, 5/9, 1/9$, and $16/9$, respectively.
\end{theorem}
Proofs are deferred to the Appendix. In addition to theoretical results on zero stability and optimal coefficients, we conduct experiments to verify whether zero stability well predicts CNNs' performance and whether the theoretically optimal coefficients work well in practice.

\section{Experiments}\label{Experiments}

In this section, we conduct extensive experiments to verify if zero stability well predicts performance with the aid of 3rd-order-discretization-based CNNs, i.e., ZeroSNets with zero stability and others without zero stability. Besides, we build a trainable version of ZeroSNet for the comparison on several benchmarks. In addition, we add different types of noise to the images and observe the relationship between robustness and zero stability. Note that hyperparameters for CIFAR-10 and CIAFR-100 are the same as those in \cite{lu2018beyond}.

\begin{table}[bt]
    \centering
      \begin{tabular}{l|c|c|c}
      \hline \hline
      \# Layers & Roots' moduli & Z. S. & Test acc. (\%)\\
      \hline \hline
      32    & 0.57, 1.00, 2.18 & No    & 77.55$\pm$0.15 \\
      32    & 1.84, 0.74, 0.74 & No    & 78.44$\pm$0.67 \\
      32    & 4.24, 0.24, 1.00 & No    & 79.67$\pm$0.49 \\
      32    & 1.88, 0.13, 1.00 & No    & 88.77$\pm$0.24 \\
      \hline
      32    & 1.00, 0.87, 0.87 & Yes & 89.44$\pm$0.08 \\
      32    & 0.94, 0.33, 0.33 & Yes & 92.59$\pm$0.15 \\
      32    & 0.81, 0.61, 0.61 & Yes & 92.84$\pm$0.10 \\
      32    & 1.00, 0.52, 0.52 & Yes & 93.14$\pm$0.07 \\
      32    & 0.60, 0.82, 0.82 & Yes & 93.27$\pm$0.04 \\
      32    & 0.33, 0.33, 1.00 & Yes & \textbf{93.28}$\pm$0.12 \\
      \hline       \hline
      44    & 0.57, 1.00, 2.18 & No    & 77.91$\pm$0.62 \\
      44    & 1.84, 0.74, 0.74 & No    & 78.70$\pm$0.43 \\
      44    & 4.24, 0.24, 1.00 & No    & 79.95$\pm$0.23 \\
      44    & 1.88, 0.13, 1.00 & No    & 83.08$\pm$0.58 \\
      \hline
      44    & 1.00, 0.87, 0.87 & Yes & 92.70$\pm$0.17 \\
      44    & 0.94, 0.33, 0.33 & Yes & 93.06$\pm$0.20 \\
      44    & 0.81, 0.61, 0.61 & Yes & 93.10$\pm$0.11 \\
      44    & 0.60, 0.82, 0.82 & Yes & 93.68$\pm$0.07 \\
      44    & 0.33, 0.33, 1.00 & Yes & 93.68$\pm$0.15 \\
      44    & 1.00, 0.52, 0.52 & Yes & \boldmath{}\textbf{93.72}$\pm$0.10\unboldmath{} \\
      \hline       \hline
      56    & 1.84, 0.74, 0.74 & No    & 78.29$\pm$0.09 \\
      56    & 0.57, 1.00, 2.18 & No    & 78.34$\pm$0.30 \\
      56    & 4.24, 0.24, 1.00 & No    & 79.39$\pm$0.17 \\
      56    & 1.88, 0.13, 1.00 & No    & 83.11$\pm$0.13 \\
      \hline
      56    & 1.00, 0.87, 0.87 & Yes & 92.42$\pm$0.35 \\
      56    & 0.81, 0.61, 0.61 & Yes & 93.08$\pm$0.36 \\
      56    & 0.94, 0.33, 0.33 & Yes & 93.15$\pm$0.16 \\
      56    & 1.00, 0.52, 0.52 & Yes & 93.71$\pm$0.24 \\
      56    & 0.60, 0.82, 0.82 & Yes & 93.99$\pm$0.06 \\
      56    & 0.33, 0.33, 1.00 & Yes & \boldmath{}\textbf{94.04}$\pm$0.12\unboldmath{} \\
      \hline \hline
      \end{tabular}%
    \caption{Test accuracies (mean $\pm$ standard deviation) obtained by setting different coefficients ($\alpha_0$, $\alpha_1$, $\alpha_2$, and $\beta$) on CIFAR-10 dataset. These coefficients give different moduli of roots. ``\# Layers'', ``Z. S.'', and ``acc.'' denote the number of layers, zero stability, and the accuracy, respectively. Similar with first- and second-order discretizations, the third-order discretization's performance can also be well predicted by zero stability. }
    \label{tab:ZeroSNetC10}%
  \end{table}%

\subsection{Predicting Performance by Zero Stability} \label{ExpZSWPP}

In early parts of this aper, preliminary experiments imply that zero stability well predicts the model performance. To further verify this conjecture, we use many 3rd-order-discretization-based CNNs for evaluations. We carefully choose coefficients $\alpha_0, \alpha_1, \alpha_2$, and $\beta$ to include more root patterns (see the Table \ref{tab:Mapping} for the mapping of those coefficients and the moduli of roots). Then, we provide Table \ref{tab:ZeroSNetC10} to show the results for 32- to 56-layer models on CIFAR-10. It is clear that if the roots satisfy the root condition (i.e., the model is zero-stable), the model performs well; If the model is non-zero-stable, its performance is relatively poor. Figure \ref{RootsAndAccC100L56} shows results of 56-layer models on CIFAR-100, and the performance gap between the zero-stable and non-zero-stable models is significant. In Fig. \ref{RootsAndAccC100L56}, the root condition (Condition \ref{rootcond}) well predicts the model performance.
Optimal coefficients given by Theorem \ref{THEOW} leads to a group of moduli of roots being $0.33, 0.33, 1.00$; These optimal coefficients are denoted in cyan hollow circles in Fig. \ref{RootsAndAccC100L56}. Combining Table \ref{tab:ZeroSNetC10} and Fig. \ref{RootsAndAccC100L56}, we find that the optimal coefficients given by Theorem \ref{THEOW} indeed outperform other coefficients in most cases. Empirically, zero stability well predicts model performance on different datasets with different discretization orders. 

\subsection{Comparison Experiments}\label{ComparExp}
\begin{table}[tb]
  \centering

    \begin{tabular}{l|c|c|c}
    \hline \hline
    Model & \# Layer & C100 (\%) & C10 (\%)\\
    \hline \hline
    ResNet & 20    & 69.46$\pm$0.15$^{*}$ & 91.25$^{\dagger}$ \\
    LM-ResNet & 20    & 69.32$\pm$0.33 & 91.67$^{\dagger}$ \\
    ZeroSNet-Opt & 20    & 69.88$\pm$0.21 & 92.01$\pm$0.35 \\
    ZeroSNet-Tra & 20    & \boldmath{}\textbf{69.90}$\pm$0.26\unboldmath{} & \boldmath{}\textbf{92.32}$\pm$0.10\unboldmath{} \\
    \hline
    ResNet & 32    & 71.30$\pm$0.20$^{*}$ & 92.49$^{\dagger}$ \\
    LM-ResNet & 32    & 71.32$\pm$0.36 & 92.82$^{\dagger}$ \\
    ZeroSNet-Opt & 32    & \boldmath{}\textbf{71.25}$\pm$0.31\unboldmath{} & \boldmath{}\textbf{93.28}$\pm$0.12\unboldmath{} \\
    ZeroSNet-Tra & 32    & 71.09$\pm$0.21 & 93.07$\pm$0.14 \\
    \hline
    ResNet & 44    & 72.36$\pm$0.23$^{*}$ & 92.83$^{\dagger}$ \\
    LM-ResNet & 44    & 72.05$\pm$0.25 & 93.34$^{\dagger}$ \\
    ZeroSNet-Opt & 44    & \boldmath{}\textbf{72.49}$\pm$0.30\unboldmath{} & 93.68$\pm$0.15 \\
    ZeroSNet-Tra & 44    & 72.18$\pm$0.38 & \boldmath{}\textbf{93.69}$\pm$0.21\unboldmath{} \\
    \hline
    ResNet & 56    & 72.56$\pm$0.08$^{*}$ & 93.03$^{\dagger}$ \\
    LM-ResNet & 56    & 72.94$\pm$0.19 & 93.69$^{\dagger}$ \\
    ZeroSNet-Opt & 56    & \boldmath{}\textbf{73.11}$\pm$0.33\unboldmath{} & \boldmath{}\textbf{94.04}$\pm$0.12\unboldmath{} \\
    ZeroSNet-Tra & 56    & 72.72$\pm$0.13 & 93.8$\pm$0.22 \\
    \hline
    ResNet & 110   & 72.24$^{*{\dagger}}$ & 93.63$^{*{\dagger}}$ \\
    LM-ResNet & 110   & 74.13$^{\dagger}$ & 93.84$^{\dagger}$ \\
    ZeroSNet-Opt & 110   & 74.50$\pm$0.28 & \boldmath{}\textbf{94.35}$\pm$0.14\unboldmath{} \\
    ZeroSNet-Tra & 110   & \boldmath{}\textbf{74.56}$\pm$0.23\unboldmath{} & 94.30$\pm$0.02 \\
    \hline \hline
    \end{tabular}%
    \caption{Test accuracies (mean $\pm$ standard deviation) on CIFAR-10 and CIFAR-100 datasets. ``\# Layer'' denotes the number of layers. Best results of each \# Layers are bold. ``$^{\dagger}$'' indicates results obtained from \cite{lu2018beyond}; ``$^*$'' indicates ResNet with the pre-activation (PreResNet). }
  \label{tab:CompC}%
\end{table}%

\begin{table}[bt]
  \centering
    \begin{tabular}{l|c|c|c}
      \hline \hline
    Model & \# Layer & Top-1 (\%) & Top-5  (\%) \\
    \hline \hline
    PreResNet & 18    & 69.66 & 88.94\\
    ZeroSNet-Opt & 18    & \textbf{69.84} & \textbf{88.97} \\
    \hline
    PreResNet & 34    & 72.21 & 90.68\\
    ZeroSNet-Opt & 34    & \textbf{72.69} & \textbf{90.83} \\
    \hline
    PreResNet & 50    & 74.51 & 91.91\\
    ZeroSNet-Opt & 50    &   \textbf{74.88}
    &  \textbf{92.03}\\
    \hline \hline
    \end{tabular}%
    \caption{Accuracies (top-1 and top-5) on ImageNet validation set with single-crop testing. ``\# Layer'' denotes the number of layers. We apply the mixed-precision training for all models on ImageNet. }
  \label{tab:CompI}%
\end{table}%

In this part, we compare the ZeroSNet with existing high-order-discretization CNNs (LM-ResNets) and PreResNets \cite{lu2018beyond, he2016identity} on CIAFR-10 and CIFAR-100 datasets. In addition, comparisons on ImageNet are also performed. Although ZeorSNet outperforms existing high-order CNNs and PreResNets, our major goal is not to beat the state-of-the-art model. Thus, we do not involve additional tricks. We provide Table \ref{tab:CompC} to show the test performance of 20- to 110-layer models on CIAFR-10 and CIFAR-100 datasets. In addition, we use ZeroSNet-Opt to represent a ZeroSNet with optimal coefficients (i.e., Theorem \ref{THEOW}) in this comparison. By setting $\lambda_n$ as a trainable parameter for the $n$th block ($n=2,3,\ldots$), we have a trainable ZeroSNet, namely ZeroSNet-Tra. Table \ref{tab:CompC} shows that ZeroSNets outperform LM-ResNets \cite{lu2018beyond} and PreResNets. Compared to ZeroSNet-Tra with several trainable $\lambda_n$, ZeroSNet-Opt with one training-free $\lambda$ shared for all blocks is competitive. In addition, Table \ref{tab:CompI} shows that the ZeroSNet-Opt has the advantage of top-1 and top-5 accuracies on ImageNet. Note that we use mixed-precision training for all models on ImageNet.

\subsection{Robustness} \label{SecRobust}
\begin{table*}[bt]
  \centering
  \begin{small}
    \begin{tabular}{l|c||c||c|c||c|c|c||c}
      \hline \hline
    Roots' moduli & Z. S. & Noise-free & $[-0.08, 0]$ & $[0, 0.08]$ & $\delta = 0.01$ & $\delta = 0.02$ & $\delta = 0.04$ & $\mu=0.3$\\
    \hline \hline
    1.84, 0.74, 0.74 & No    & 78.29$\pm$0.09 & 67.01$\pm$1.08 & 67.14$\pm$1.32 & 77.17$\pm$0.28 & 70.65$\pm$0.78 & 50.76$\pm$3.94 & 66.82$\pm$1.36 \\
    0.57, 1.00, 2.18 & No    & 78.34$\pm$0.30 & 66.16$\pm$1.81 & 66.99$\pm$1.67 & 76.88$\pm$0.48 & 69.75$\pm$1.38 & 48.72$\pm$2.82 & 67.81$\pm$0.22 \\
    4.24, 0.24, 1.00 & No    & 79.39$\pm$0.17 & 67.06$\pm$1.83 & 67.06$\pm$2.02 & 77.41$\pm$0.66 & 70.19$\pm$1.84 & 51.23$\pm$0.39 & 69.31$\pm$0.97 \\
    1.88, 0.13, 1.00 & No    & 83.11$\pm$0.13 & 69.38$\pm$0.40 & 69.16$\pm$0.51 & 81.25$\pm$0.22 & 73.23$\pm$0.37 & 50.2$\pm$0.96 & 73.88$\pm$0.96\\
    \hline
    1.00, 0.87, 0.87 & Yes   & 92.42$\pm$0.35 & 83.89$\pm$1.55 & 83.98$\pm$0.81 & 91.21$\pm$0.43 & 86.73$\pm$0.94 & 64.69$\pm$2.35 & 86.75$\pm$0.77 \\
    0.81, 0.61, 0.61 & Yes   & 93.08$\pm$0.36 & 86.91$\pm$0.31 & 86.92$\pm$0.29 & 92.00$\pm$0.28 & 88.66$\pm$0.30 & 74.19$\pm$0.39 & 87.43$\pm$0.48 \\
    0.94, 0.33, 0.33 & Yes   & 93.15$\pm$0.16 & 88.05$\pm$0.23 & \boldmath{}\textbf{88.16}$\pm$0.29\unboldmath{} & 92.24$\pm$0.18 & 89.51$\pm$0.19 & \boldmath{}\textbf{76.74}$\pm$0.13\unboldmath{} & 87.73$\pm$0.32 \\
    1.00, 0.52, 0.52 & Yes   & 93.71$\pm$0.24 & 87.10$\pm$0.38 & 87.23$\pm$0.27 & 92.47$\pm$0.30 & 88.84$\pm$0.29 & 73.96$\pm$0.47 & \boldmath{}\textbf{88.90}$\pm$0.35\unboldmath{} \\
    0.60, 0.82, 0.82 & Yes   & 93.99$\pm$0.06 & 87.76$\pm$0.31 & 87.84$\pm$0.42 & 92.75$\pm$0.14 & 89.46$\pm$0.29 & 74.97$\pm$0.84 & 88.82$\pm$0.29\\
    0.33, 0.33, 1.00 & Yes   & \boldmath{}\textbf{94.04}$\pm$0.12\unboldmath{} & \boldmath{}\textbf{88.31}$\pm$0.33\unboldmath{} & 87.64$\pm$0.43 & \boldmath{}\textbf{92.96}$\pm$0.12\unboldmath{} & \boldmath{}\textbf{89.67}$\pm$0.29\unboldmath{} & 74.49$\pm$1.25 &88.78$\pm$0.29\\
    \hline \hline
    \end{tabular}%
  \end{small}
    \caption{Test accuracies (mean $\pm$ standard deviation) on CIFAR-10 under uniform noise ($[$lower bound, upper bound$]$), zero-mean Gaussian noise (with standard deviation $\delta$), and constant noise (with magnitude $\mu$). Note the input imgages are normalized into an interval of $[0,1]$. ``Z. S.'' denotes zero stability. }    
    \label{tab:Rob}%
\end{table*}%
We verify the robustness of models on the test set. We store network parameters after the noise-free training. Then, we unnormalize the input images into $[0, 1]$. After feeding these input images into the stored models, the accuracies under perturbations are obtained. Three types of noises are involved: Uniform noise, Gaussian noise, and constant noise. 
Each type of noise is added to input images with different levels. Table \ref{tab:Rob} shows the test performance of 56-layer models under these three types of noises. The uniform noises are in $[$lower bound, upper bound$]$; The Gaussian noises are generated with standard deviation $\delta$ and a mean of zero; The constant noise is a grey image with pixel values of $\mu$. As in Table \ref{tab:Rob}, the non-zero-stable models' test accuracies decrease dramatically after injecting noises, while zero-stable models are robust. For example, under uniform noises distributed in $[-0.08, 0]$, test accuracies of non-zero-stable models decrease $12.38\%$ on average, while for zero-stable models, this degradation is only $\boldsymbol{6.40\%}$. Similar phenomenons are clear in other noise-model pairs. We provide more experiment results on noises with different levels and some results with adversarial examples (i.e., fast gradient sign method (FGSM) \cite{goodfellow2014explaining} on MNIST and projected gradient descent (PGD) \cite{madry2018towards} on CIFAR-10) in the Appendix.

\subsection{Generalization Gap}
In addition to performance on the test set, we provide the experimental results of the generalization gap for ZeroSNets in Table \ref{tab:GeneralizationC100}. To facilitate comparison, all involved ZeroSNets have a root as $1$ and two repeated roots. From tables \ref{tab:GeneralizationC100}, we can see that smaller moduli of roots (which imply better zero stability) generally lead to a smaller generalization gap. In general, the optimal coefficients given by Theorem \ref{THEOW} lead to the best generalization gap. To achieve sufficient training, we train all models for $500$ epochs for generalization gap experiments (this is different from all other experiments in this paper).  
\begin{table}[tb]
  \centering
  \begin{small}
    \begin{tabular}{l||c|c|c||c}
      \hline
      \hline
    Model & Roots' moduli  & Training & Test  & Gap \\
    \hline    \hline
    ZeroSNet44 & 1, 0.52, 0.52 & 99.31 & 72.25 & 27.06 \\
    ZeroSNet44 & 1, 0.87, 0.87 & 98.04 & 71.59 & 26.45 \\
    ZeroSNet44 & 1, 0.33, 0.33 & 99.34 & 72.91 & \textbf{26.43} \\
    \hline
    ZeroSNet56 & 1, 0.87, 0.87 & 99.31 & 71.07 & 28.24 \\
    ZeroSNet56 & 1, 0.52, 0.52 & 99.67 & 72.55 & 27.12 \\
    ZeroSNet56 & 1, 0.33, 0.33 & 99.65 & 72.96 & \textbf{26.69} \\
    \hline
    ZeroSNet110 & 1, 0.87, 0.87 & 99.90 & 73.62 & 26.28 \\
    ZeroSNet110 & 1, 0.52, 0.52 & 99.90 & 75.15 & \textbf{24.75} \\
    ZeroSNet110 & 1, 0.33, 0.33 & 99.92 & 75.00 & 24.92 \\
    \hline
    ZeroSNet164 & 1, 0.87, 0.87 & 98.77 & 73.24 & 25.53 \\
    ZeroSNet164 & 1, 0.52, 0.52 & 99.92 & 77.77 & 22.15 \\
    ZeroSNet164 & 1, 0.33, 0.33 & 99.92 & 78.15 & \textbf{21.77} \\
    \hline
    ZeroSNet326 & 1, 0.87, 0.87 & 99.47 & 73.38 & 26.09 \\
    ZeroSNet326 & 1, 0.52, 0.52 & 99.95 & 78.63 & 21.32 \\
    ZeroSNet326 & 1, 0.33, 0.33 & 99.95 & 79.26 & \textbf{20.69} \\
    \hline    \hline
    \end{tabular}%
  \end{small}
    \caption{Generalization gap (\%) on CIFAR-100. We use difference of the training and test accuracies (i.e., ``training acc. $-$ test acc.'') to measure the generalization gap. Generally, as the moduli of roots decrease, generalization ability of the corresponding model improves. }
  \label{tab:GeneralizationC100}%
\end{table}%

\subsection{Computation Efficiency} \label{sec:Efficiency}

From the Experiment section, we can see that there are performance improvements brought by ZeroSNets. In this part, we evaluate the costs of such improvements. 
A comparison of the number of parameters is given in Table \ref{tab:Parameters}. ZeroSNets have a close number of parameters compared with PreResNets. Besides, we provide the runtime of PreResNet20 and ZeroSNet20 on CIFAR-10 (Table \ref{tab:time}). Table \ref{tab:time} shows that time consumption of ZeroSNet20 is close to PreResNet20, especially for large batch sizes. When we perform the runtime experiments, we remain only one task on a server. 

\begin{table}[bt]
  \centering
  \begin{small}
    \begin{tabular}{l||c|c|c}
      \hline  \hline
    \# Layers  & LMResNet & PreResNet & ZeroSNet\\
    \hline \hline
    20     & 0.27M & 0.28M & 0.28M \\
    \hline
    32     & 0.47M & 0.48M & 0.48M \\
    \hline
    44     & 0.66M & 0.67M & 0.67M \\
    \hline
    56     & 0.86M & 0.87M & 0.87M \\
    \hline
    110    & 1.14M & 1.74M  & 1.74M \\

    \hline \hline
    \end{tabular}%
\end{small}
    \caption{Parameter amount of ResNets, LMResNets, PreResNets, and ZeroSNets. Note that the parameter amount of the ZeroSNet is close to PreResNet. }
  \label{tab:Parameters}%
\end{table}%

\begin{table}[bt]
  \centering
  \begin{small}
  \begin{tabular}{l||c|c|c}
    \hline  \hline
  Model                  & BS=128 & BS=512 & BS=4096 \\
  \hline  \hline
  PreResNet20 (training) & 1532   & 829       & 913     \\ \hline
  ZeroSNet20 (training)  & 1662  & 829       & 928     \\ \hline \hline
  PreResNet20 (test)     & 142    & 140         & 212 \\ 
  \hline
  ZeroSNet20 (test)      & 151   & 143       & 220     \\ 
  \hline \hline
  \end{tabular}
  \end{small}
  \caption{Training and test runtime (second) of PreResNet20 and ZeroSNet20 on CIFAR-10. }
  \label{tab:time}
\end{table}

\subsection{Experiment Settings} \label{SecExpSet}
  
We provide detailed experiment settings as follows. We use Pytorch 1.8.1 framework and run our experiments on a server with 10 RTX 2080 TI GPUs and 2 RTX 3090 GPUs. 

  \textbf{CIFAR.}
  Hyperparameters for CIFAR-10 and CIAFR-100 are the same as those in \cite{lu2018beyond}. We conduct all experiments with stochastic gradient descent (SGD) optimizer. On the CIFAR, we use a batch size of $128$ with an initial learning rate of $0.1$, the momentum of $0.9$, and weight decay $0.0001$. Models in generalization gap experiments (Table \ref{tab:GeneralizationC100}) are trained for $500$ epochs to achieve sufficient training. Except for the generalization gap experiments, all models on CIFAR-$10$ and CIFAR-$100$ are trained for $160$ and $300$ epochs, respectively. We apply the step decay to train all models on CIFAR and divide the learning rate by $10$ at half and three-quarters of the total epoch. We report the ``mean $\pm$ standard deviations'' accuracies based on three individual runs. For the trainable version of ZeroSNet (i.e., ZeroSNet-Tra), all $\lambda_n$ are initialized as $1$. The data augmentations are the random crop with a $4$-pixel padding and random horizontal flip, as in \cite{lu2018beyond}. 
  
  \textbf{ImageNet.}
  Our training script is based on \url{https://github.com/13952522076/Efficient_ImageNet_Classification} and remains all default hyperparameters. To improve the training efficiency on ImageNet, we use a mix-precision strategy provided by NVIDIA apex with distributed training. We apply the cosine decay with a $5$-epoch warmup to train models for $150$ epochs. The weight decay and the momentum are $ 4\times10^{-5}$ and $0.9$, respectively. Following the adjustment guidance of the learning rate and the batch size \cite{goyal2017accurate, jastrzkebski2018width}, we set them according to the GPU memory. Specifically, for $18$-layer models, we use an initial learning rate of $0.2$ and a batch size of $128$; for $34$-layer models, we use an initial learning rate of $0.1$ and a batch size of $64$; for $50$-layer models, we use an initial learning rate of $0.05$ and a batch size of $32$. For ImageNet, we apply 8-GPU distributed training on a single server.
  
  \textbf{Robustness.}
  The random seeds of PyTorch for generating the uniform and Gaussian noises are both 1. In the standard training phase, we store three individual models for each group of $\alpha_0$, $\alpha_1$, $\alpha_2$, and $\beta$. Then, we use the three models to evaluate the average robustness and report the result in the ``mean $\pm$ standard deviations'' format. Finally, we map the pixels from $[0, 255]$ to $[0, 1]$. After noise injection, we clip the dirty data (negative-valued input pixels or values exceed $1$) within $[0, 1]$.

  Table \ref{tab:Mapping} gives the mapping from coefficients to moduli of roots. 
  \begin{table*}[]
    \centering
    \begin{small}
      \begin{tabular}{c|c|c|c||c|c|c||c}
        \hline \hline
      $\alpha_0$ & $\alpha_1$ & $\alpha_2$ & $\beta_0$ & Module of the 1st root & Module of the 2nd root & Module of the 3rd root & Z. S. \\
      \hline \hline
      1.0000  & 1.0000  & 1.0000  & 1.0000  & 1.84  & 0.74  & 0.74  & No \\
      3.7500  & -4.0000  & 1.2500  & -0.5000  & 0.57  & 1.00  & 2.18  & No \\
      -3.0000  & 5.0000  & -1.0000  & 4.0000  & 4.24  & 0.24  & 1.00  & No \\
      -0.7500  & 2.0000  & -0.2500  & 2.5000  & 1.88  & 0.13  & 1.00  & No \\
      \hline
      2.2500  & -2.0000  & 0.7500  & 0.5000  & 1.00  & 0.87  & 0.87  & Yes \\
      0.1000  & 0.2000  & 0.3000  & 0.4000  & 0.81  & 0.61  & 0.61  & Yes \\
      0.5000  & 0.3000  & 0.1000  & 0.1000  & 0.94  & 0.33  & 0.33  & Yes \\
      0.8250  & -0.1000  & 0.2750  & 1.4500  & 1.00  & 0.52  & 0.52  & Yes \\
      1.0000  & 0.3000  & -0.4000  & 1.0000  & 0.60  & 0.82  & 0.82  & Yes \\
      0.3333  & 0.5556  & 0.1111  & 1.7778  & 0.33  & 0.33  & 1.00  & Yes \\
      \hline \hline
      \end{tabular}%
    \end{small}
      \caption{Mapping from coefficients to moduli of roots. ``Z. S.'' denotes zero stability. Note that the theoretically optimal coefficients (1/3, 5/9, 1/9, and 16/9) are in decimal forms here (0.3333, 0.5556, 0.1111, and 1.7778).
      }
    \label{tab:Mapping}%
  \end{table*}%

\section{Related Work}\label{rw}


\textbf{Robustness of neural ODEs:}
\citeauthor{hanshu2019robustness} gives a loss term to minimize the upper bound of the difference between end states and find that neural ODEs with continuous representation perform well on the robustness. \citeauthor{zhang2019towards} study the robustness through the lens of step size, and they find that small step size benefits both forward and backward propagation. Embedding Gaussian processes into a neural ODE improves the robustness, as in \cite{anumasa2021improving}. By training multiple noise-injected ResNets to approximate the Feynman-Kac formula, a robust model is constructed in \cite{wang2019resnets}. Differently, we consider the robustness of discrete CNNs and bridge it with the network structure through zero stability. 

\textbf{Stability of CNNs:} A-stability of CNNs is investigated in \cite{haber2017stable}. The insight that the features should be well-posed in \cite{haber2017stable} is important for keeping the representation ability and away from explosions. Although the generalization is mentioned, the connection between it and A-stability is not clear in \cite{haber2017stable}. Since A-stability does not involve perturbation, it may be irrelevant to the generalization. \cite{weinan2017proposal, lu2018beyond, chen2019ordinary} give the interpretation of deep neural networks from an ordinary differential equation (ODE) perspective. Based on those works, \cite{ruthotto2020deep} further studies stability from a perspective of the partial differential equation (PDE). \cite{ruthotto2020deep} constructs parabolic and hyperbolic CNNs, and proves that under certain assumptions (e.g., weight symmetry, special activation), the parabolic and hyperbolic CNNs are stable.  
Different from \cite{ruthotto2020deep}, we construct zero-stable CNNs based on high-order discretization and show that zero stability can predict performance well. \cite{zhang2020forward} studies the stability of several ResNet-like networks, and it gives upper bounds of the output feature maps and the sensitivity bound. Differently, we use zero stability in numerical analysis \cite{gautschi1997numerical} and then provide guidance to construct high-order structures.

\textbf{Structure based on high-order discretization:} After interpreting some well-performed CNNs as ODEs, \citeauthor{lu2018beyond} give the LM architecture. We interpret the LM architecture as a second-order discretization and use it as a tool for our preliminary observation on how zero stability affects model performance. Unlike the LM architecture, ZeroSNet in our work has a theoretical guarantee to be consistent and zero-stable. In our experiments, following the same settings of hyperparameters, ZeroSNet outperforms the LM-ResNet in \cite{lu2018beyond}. 



\section{Discussion}\label{Discussion}
The well-performed ZeroSNet is somehow just a by-product for investigating the nature of CNNs. To speed up the training, we use plain settings for all experiments and apply the mixed-precision training on ImageNet, and our results cannot beat the state-of-the-art ones on the leaderboard. Besides, due to the space limitation, we only discuss the first- to third-order discretizations, but we believe the connection between performance and zero-stability is clear. A general theory for leading the structure designing is beyond this paper's scope, and it requires further exploration.
The precise understanding of deep neural networks still needs more effort, and our work only takes a little step to this big problem's answer.

\section{Conclusion}\label{Conclusion}

In this work, we first observe that zero stability well predicts the performance of PreResNets and LM-ResNets. Based on these preliminary observations, we construct a high-order CNN named ZeroSNet to further verify the prediction ability of zero stability. Theoretically, we prove ZeroSNet's advantages on consistency and zero-stability, with a group of optimal coefficients for historical features and the current activations deduced. Four groups of experiments are carried out in this paper. First, we compare ZeroSNets with their non-zero-stable counterparts, and the results clearly show that zero-stable models outperform non-zero-stable ones on generalization. Second, we evaluate the theoretically optimal coefficients on different datasets, and the results demonstrate that they are also optimal in practice. Then, ZeroSNet with the theoretically optimal coefficients and ZeroSNet with trainable parameters are employed for comparison. Results show that ZeroSNets outperform previous advanced CNNs on CIFAR-10, CIFAR-100, and ImageNet. Finally, experiments on test images injected with noise verify the superiority of zero-stable CNNs on the robustness. 

\setcounter{secnumdepth}{0} 
\section{Acknowledgements}\label{Acknowledgements}
This work was supported in part by the CAS ``Light of West China'' Program, in part by the Natural Science Foundation of Chongqing (China) under Grant cstc2020jcyj-zdxmX0028, in part by the National Nature Science Foundation of China under Grant 62176109, Grant 62072429, and Grant 61902370, in part by the Key Cooperation Project of Chongqing Municipal Education Commission under Grant HZ2021017 and Grant HZ2021008, and in part by the Chongqing Entrepreneurship and Innovation Support Program for Overseas Returnees under Grant CX2021100. 
\normalem
\bibliography{Reference}

\newpage
\part*{\centering Appendix}
\appendix

\setcounter{secnumdepth}{2} 

\section{Additional Experiments}
We provide more robustness results with uniform noises, Gaussian noises, and constant noises injection in Tables \ref{tab:RandTest}, \ref{tab:RandnTest}, and \ref{tab:ConstTest}, respectively. The results consistently show that zero-stable models outperform the non-zero-stable counterparts. All results in these three tables are obtained with 56-layer third-order-discretization-based CNNs on CIFAR-10. 

Besides, we also provide robustness results with FGSM and PGD adversarial examples. Tables \ref{tab:PGDC10} and \ref{tab:FGSMMNIST} show the robustness results with PGD on CIFAR-10 and with FGSM on MNIST, respectively. On on CIFAR-10, we follow settings in \url{https://github.com/AlbertMillan/adversarial-training-pytorch} and train these models for 80 epochs with BS=128, LR=0.1 (multiplied by 0.1 at epoch 75), momentum=0.9, and weight decay=2e-4. On MNIST, we follow settings in \url{https://github.com/1Konny/FGSM} and train these models for 20 epochs with BS=100, LR=2e-4. It is clear that the zero-stable models' robustness under FGSM and PGD perturbations is better than the non-zero-stable counterparts.



  \begin{table*}[]
    \centering
      \begin{tabular}{l|c|c|c|c|c|c|c}
      \hline \hline
      Roots' moduli & Z. S. & $[-0.08, 0]$ & $[-0.06, 0]$ & $[-0.04, 0]$ & $[0, 0.04]$ & $[0, 0.06]$ & $[0, 0.08]$\\
      \hline \hline
      1.84, 0.74, 0.74 & No    & 67.01$\pm$1.08 & 72.34$\pm$0.63 & 76.12$\pm$0.52 & 76.39$\pm$0.46 & 72.67$\pm$0.73 & 67.14$\pm$1.32 \\
      0.57, 1.00, 2.18 & No    & 66.16$\pm$1.81 & 71.94$\pm$1.02 & 75.96$\pm$0.73 & 76.33$\pm$0.7 & 72.4$\pm$1.07 & 66.99$\pm$1.67 \\
      4.24, 0.24, 1.00 & No    & 67.06$\pm$1.83 & 72.48$\pm$1.55 & 76.55$\pm$0.84 & 76.66$\pm$0.87 & 72.35$\pm$1.7 & 67.06$\pm$2.02 \\
      1.88, 0.13, 1.00 & No    & 69.38$\pm$0.40 & 75.85$\pm$0.39 & 80.13$\pm$0.23 & 80.17$\pm$0.10 & 75.8$\pm$0.05 & 69.16$\pm$0.51 \\       \hline
      1.0, 0.87, 0.87 & Yes   & 83.89$\pm$1.55 & 88.28$\pm$1.01 & 90.81$\pm$0.53 & 90.87$\pm$0.31 & 88.06$\pm$0.60 & 83.98$\pm$0.81 \\
      0.81, 0.61, 0.61 & Yes   & 86.91$\pm$0.31 & 89.73$\pm$0.22 & 91.54$\pm$0.24 & 91.63$\pm$0.33 & 89.7$\pm$0.06 & 86.92$\pm$0.29 \\
      0.94, 0.33, 0.33 & Yes   & 88.05$\pm$0.23 & 90.38$\pm$0.22 & 91.94$\pm$0.13 & 91.96$\pm$0.09 & 90.43$\pm$0.12 & \boldmath{}\textbf{88.16}$\pm$0.29\unboldmath{} \\
      1.0, 0.52, 0.52 & Yes   & 87.10$\pm$0.38 & 90.13$\pm$0.17 & 92.14$\pm$0.28 & 92.12$\pm$0.26 & 90.17$\pm$0.26 & 87.23$\pm$0.27 \\
      0.6, 0.82, 0.82 & Yes   & 87.76$\pm$0.31 & 90.53$\pm$0.30 & 92.46$\pm$0.21 & 92.56$\pm$0.10 & 90.52$\pm$0.34 & 87.84$\pm$0.42 \\
      0.33, 0.33, 1.00 & Yes   & \boldmath{}\textbf{88.31}$\pm$0.33\unboldmath{} & \boldmath{}\textbf{90.84}$\pm$0.22\unboldmath{} & \boldmath{}\textbf{92.57}$\pm$0.07\unboldmath{} & \boldmath{}\textbf{92.53}$\pm$0.02\unboldmath{} & \boldmath{}\textbf{90.58}$\pm$0.12\unboldmath{} & 87.64$\pm$0.43\\
      \hline \hline
      \end{tabular}%
      \caption{Test accuracies (mean $\pm$ standard deviation) under uniform noise ($[$lower bound, upper bound$]$) injections. ``Z. S.'' denotes zero stability. }
    \label{tab:RandTest}%
  \end{table*}%

  \begin{table*}[]
    \centering
      \begin{tabular}{l|c|c|c|c|c|c|c}
      \hline \hline
      Roots' moduli & Z. S. & $\delta = 0$ & $\delta = 0.01$ & $\delta = 0.02$ & $\delta = 0.03$ & $\delta = 0.035$ & $\delta = 0.04$ \\
      \hline \hline
      1.84, 0.74, 0.74 & No & 78.29$\pm$0.09    & 77.17$\pm$0.28 & 70.65$\pm$0.78 & 60.51$\pm$2.56 & 55.45$\pm$3.27 & 50.76$\pm$3.94 \\
      0.57, 1.00, 2.18 & No & 78.34$\pm$0.30   & 76.88$\pm$0.48 & 69.75$\pm$1.38 & 59.09$\pm$2.50 & 53.79$\pm$2.50 & 48.72$\pm$2.82 \\
      4.24, 0.24, 1.00 & No & 79.39$\pm$0.17   & 77.41$\pm$0.66 & 70.19$\pm$1.84 & 60.49$\pm$1.22 & 55.52$\pm$0.74 & 51.23$\pm$0.39 \\
      1.88, 0.13, 1.00 & No & 83.11$\pm$0.13   & 81.25$\pm$0.22 & 73.23$\pm$0.37 & 61.44$\pm$0.51 & 55.54$\pm$0.95 & 50.20$\pm$0.96 \\       \hline
      1.00, 0.87, 0.87 & Yes & 92.42$\pm$0.35  & 91.21$\pm$0.43 & 86.73$\pm$0.94 & 77.57$\pm$1.93 & 71.39$\pm$2.18 & 64.69$\pm$2.35 \\
      0.81, 0.61, 0.61 & Yes & 93.08$\pm$0.36  & 92.00$\pm$0.28 & 88.66$\pm$0.30 & 82.54$\pm$0.36 & 78.60$\pm$0.39 & 74.19$\pm$0.39 \\
      0.94, 0.33, 0.33 & Yes & 93.15$\pm$0.16  & 92.24$\pm$0.18 & 89.51$\pm$0.19 & \boldmath{}\textbf{84.4}$\pm$0.18\unboldmath{} & \boldmath{}\textbf{80.85}$\pm$0.04\unboldmath{} & \boldmath{}\textbf{76.74}$\pm$0.13\unboldmath{} \\
      1.00, 0.52, 0.52 & Yes & 93.71$\pm$0.24  & 92.47$\pm$0.30 & 88.84$\pm$0.29 & 82.62$\pm$0.32 & 78.43$\pm$0.33 & 73.96$\pm$0.47 \\
      0.60, 0.82, 0.82 & Yes & 93.99$\pm$0.06  & 92.75$\pm$0.14 & 89.46$\pm$0.29 & 83.75$\pm$0.51 & 79.57$\pm$0.74 & 74.97$\pm$0.84 \\
      0.33, 0.33, 1.00 & Yes & \boldmath{}\textbf{94.04}$\pm$0.12\unboldmath{}  & \boldmath{}\textbf{92.96}$\pm$0.12\unboldmath{} & \boldmath{}\textbf{89.67}$\pm$0.29\unboldmath{} & 83.46$\pm$0.57 & 79.31$\pm$0.73 & 74.49$\pm$1.25 \\
      \hline \hline
      \end{tabular}%
      \caption{Test accuracies (mean $\pm$ standard deviation) under Gaussian noise (with standard deviation $\delta$) injections. ``Z. S.'' denotes zero stability.}
    \label{tab:RandnTest}%
  \end{table*}%
\begin{table*}[]
    \centering
      \begin{tabular}{l|c|c|c|c|c|c|c}
      \hline \hline
      Roots' moduli & Z. S. & $\mu=-0.4$ & $\mu=-0.3$ & $\mu=-0.2$ & $\mu=0.2$ & $\mu=0.3$ & $\mu=0.4$ \\
      \hline \hline
      1.84,0.74,0.74 & No    & 53.98$\pm$5.12 & 64.98$\pm$3.18 & 72.20$\pm$1.54 & 73.1$\pm$0.58 & 66.82$\pm$1.36 & 56.76$\pm$2.07 \\
      0.57,1.00,2.18 & No    & 52.41$\pm$2.51 & 64.55$\pm$1.45 & 72.03$\pm$0.74 & 73.92$\pm$0.22 & 67.81$\pm$0.22 & 58.70$\pm$0.24 \\
      4.24,0.24,1.00 & No    & 52.97$\pm$3.33 & 65.05$\pm$2.44 & 73.08$\pm$1.55 & 75.06$\pm$0.72 & 69.31$\pm$0.97 & 60.16$\pm$0.83 \\
      1.88,0.13,1.00 & No    & 57.75$\pm$1.89 & 69.64$\pm$1.29 & 77.55$\pm$0.68 & 78.81$\pm$0.25 & 73.88$\pm$0.96 & 66.48$\pm$1.44 \\       \hline
      1.00,0.87,0.87 & Yes   & 72.06$\pm$1.08 & 83.33$\pm$0.95 & 89.14$\pm$0.49 & 90.01$\pm$0.41 & 86.75$\pm$0.77 & 81.27$\pm$0.81 \\
      0.81,0.61,0.61 & Yes   & 72.12$\pm$2.01 & 83.44$\pm$1.18 & 89.73$\pm$0.38 & 90.75$\pm$0.16 & 87.43$\pm$0.48 & 81.12$\pm$0.95 \\
      0.94,0.33,0.33 & Yes   & 73.03$\pm$0.79 & 84.04$\pm$0.42 & 89.82$\pm$0.13 & 90.89$\pm$0.23 & 87.73$\pm$0.32 & 82.39$\pm$0.29 \\
      1.00,0.52,0.52 & Yes   & \boldmath{}\textbf{76.57}$\pm$0.79\unboldmath{} & \boldmath{}\textbf{86.02}$\pm$0.49\unboldmath{} & \boldmath{}\textbf{91.05}$\pm$0.16\unboldmath{} & 91.68$\pm$0.25 & \boldmath{}\textbf{88.90}$\pm$0.35\unboldmath{} & \boldmath{}\textbf{83.45}$\pm$0.40\unboldmath{} \\
      0.60,0.82,0.82 & Yes   & 75.52$\pm$0.56 & 85.84$\pm$0.17 & 91.03$\pm$0.04 & 91.72$\pm$0.36 & 88.82$\pm$0.29 & 83.19$\pm$0.77 \\
      0.33,0.33,1.00 & Yes   & 74.18$\pm$1.07 & 85.31$\pm$0.64 & 90.98$\pm$0.17 & \boldmath{}\textbf{91.82}$\pm$0.06\unboldmath{} & 88.78$\pm$0.29 & 83.35$\pm$0.42 \\
      \hline \hline
      \end{tabular}%
      \caption{Test accuracies (mean $\pm$ standard deviation) under constant noise (with magnitude $\mu$) injections. ``Z. S.'' denotes zero stability. }
    \label{tab:ConstTest}%
  \end{table*}%

  \begin{table*}[hbt]
    \centering
    \begin{tabular}{l|c|c|c|c|c}
      \hline \hline
    Roots' moduli & Z. S. & $\epsilon$=0 & $\epsilon$=3/255 & $\epsilon$=5/255 & $\epsilon$=8/255 \\
    \hline \hline 
    0.57,   1.00, 2.18 & No  & 81.83 & 54.66 (-27.17) & 51.57 (-30.27) & 48.80 (-33.04) \\
    1.88,   0.13, 1.00 & No  & 88.87 & 66.05 (-22.82) & 61.74 (-27.14) & 57.11 (-31.77) \\
    \hline
    1.00,   0.52, 0.52 & Yes & 91.65 & 71.40 (-20.26) & 66.58 (-25.08) & 61.27 (-30.38)\\ 
    0.33,   0.33, 1.00 & Yes & 91.83 & \textbf{73.46 (-18.38)} & \textbf{69.33 (-22.50)} & \textbf{63.76 (-28.07)} \\
    \hline \hline
    \end{tabular}
    \caption{Adversarial accuracies (-reduction) on CIFAR-10 with PGD adversarial examples. } 
    \label{tab:PGDC10}
    \end{table*}
  
    \begin{table}[htb]
      \centering
      \begin{small}
      \begin{tabular}{l|c|c|c|c|c}
        \hline \hline
      Moduli   of roots  & Z. S. & $\epsilon$=0 & $\epsilon$=0.15 & $\epsilon$=0.3 & $\epsilon$=0.5 \\
      \hline \hline
      1.84,   0.74, 0.74 & No       & 100        & 31              & 12             & 14             \\
      0.57,   1.00, 2.18 & No       & 99         & 22              & 8              & 7              \\
      4.24,   0.24, 1.00 & No       & 100        & 34              & 22             & 17             \\
      1.88,   0.13, 1.00 & No       & 100        & 44              & 12             & 13             \\
      \hline
      1.00,   0.87, 0.87 & Yes      & 100        & 49              & 24             & 19             \\
      0.81,   0.61, 0.61 & Yes      & 100        & 76              & 42             & 19             \\
      0.94,   0.33, 0.33 & Yes      & 99         & 57              & 21             & 18             \\
      1.00,   0.52, 0.52 & Yes      & 99         & 67              & 53             & 29             \\
      0.60,   0.82, 0.82 & Yes      & 100        & 54              & 25             & 23             \\
      0.33,   0.33, 1.00 & Yes      & 99         & \textbf{83}              & \textbf{59}             & \textbf{39}           \\ \hline \hline 
      \end{tabular}
    \end{small}
      \caption{Adversarial accuracies on MNIST with FGSM adversarial examples. }
      \label{tab:FGSMMNIST}
      \end{table}
  
\section{Derivation of ZeroSNet}
We deduce ZeroSNet's formal description as follows. Taking the Taylor expansion of $\boldsymbol{y}(t_{n-1})$ at $\boldsymbol{y}(t_{n})$ yileds
\begin{equation}
  \begin{small}
    \begin{aligned}
    \boldsymbol{y} \left(t_{n-1}\right)= &\boldsymbol{y} \left(t_{n}\right)-\dot{\boldsymbol{y}}\left(t_{n}\right) (t_{n-1}- t_n )+\frac{\ddot{\boldsymbol{y}}\left(t_{n}\right)}{2!} (t_{n-1}- t_n )^{2}\\
    &-\frac{\dddot{\boldsymbol{y}}\left(\tau_{1}\right)}{3!} (t_{n-1}- t_n )^{3}.
\end{aligned}
\end{small}
\end{equation}
Similarly, we have
\begin{equation}
  \begin{small}
    \begin{aligned}
    \boldsymbol{y} \left(t_{n-2}\right)=&\boldsymbol{y} \left(t_{n}\right)-\dot{\boldsymbol{y}}\left(t_{n}\right) (t_{n-2}- t_n )+\frac{\ddot{\boldsymbol{y}}\left(t_{n}\right)}{2!} (t_{n-2}- t_n )^{2}\\
    &-\frac{\dddot{\boldsymbol{y}}\left(\tau_{2}\right)}{3!} (t_{n-2}- t_n )^{3}
\end{aligned}
\end{small}
\end{equation}
and
\begin{equation}
  \begin{small}
    \begin{aligned}
    \boldsymbol{y} \left(t_{n+1}\right)=&\boldsymbol{y} \left(t_{n}\right)-\dot{\boldsymbol{y}}\left(t_{n}\right) (t_{n+1}- t_n )+\frac{\ddot{\boldsymbol{y}}\left(t_{n}\right)}{2!} (t_{n+1}- t_n )^{2}\\
    &-\frac{\dddot{\boldsymbol{y}}\left(\tau_{3}\right)}{3!} (t_{n+1}- t_n )^{3}.
    \end{aligned}
  \end{small}
\end{equation}
Take two arbitrary coefficients $\lambda_1$ and $\lambda_2$ which satisfy $3 \lambda_2 \neq \lambda_1$, $\lambda_1 \neq 0$, and $\lambda_2 \neq 0$. Combining with the equal-interval sampling assumption, we have
\begin{equation}\label{ddy1}
    \lambda_1 \ddot{ \boldsymbol{y} }\left(t_{n}\right)=\lambda_1 (\frac{2}{h^{2}}  \boldsymbol{y} \left(t_{n-1}\right)-\frac{2}{h^{2}}  \boldsymbol{y} \left(t_{n}\right)+\frac{2}{h} \dot{ \boldsymbol{y} }\left(t_{n}\right)+\frac{h}{3} \dddot{\boldsymbol{y}} \left(\tau_{1}\right)),
\end{equation}
\begin{equation}\label{ddy2}
    \begin{aligned}
    -(\lambda_1+\lambda_2) \ddot{ \boldsymbol{y} }\left(t_{n}\right)=&-(\lambda_1+\lambda_2) (\frac{1}{2h^{2}}  \boldsymbol{y} \left(t_{n-2}\right)\\
    &-\frac{1}{2h^{2}}  \boldsymbol{y} \left(t_{n}\right)+\frac{1}{h} \dot{ \boldsymbol{y} }\left(t_{n}\right)+\frac{2h}{3} \dddot{\boldsymbol{y}} \left(\tau_{2}\right)),
\end{aligned}
\end{equation}
and
\begin{equation}\label{ddy3}
    \lambda_2 \ddot{ \boldsymbol{y} }\left(t_{n}\right)= \lambda_2 (\frac{2}{h^{2}}  \boldsymbol{y} \left(t_{n+1}\right)-\frac{2}{h^{2}}  \boldsymbol{y} \left(t_{n}\right)-\frac{2}{h} \dot{ \boldsymbol{y} }\left(t_{n}\right)-\frac{h}{3} \dddot{\boldsymbol{y}} \left(\tau_{3}\right)).
\end{equation}
Summing both sides of equations \eqref{ddy1} through \eqref{ddy3} and using $\mathcal{O}(h^2)$ to represent third-order terms yield
\begin{equation}\label{EQdoty1}
\begin{aligned}
\dot{ \boldsymbol{y} }\left(t_{n}\right)=& \frac{2 \lambda_2  }{(3 \lambda_2 -\lambda_1) h}\boldsymbol{y} \left(t_{n+1}\right)-\frac{3 (\lambda_2 +\lambda_1)}{2(3 \lambda_2 -\lambda_1) h}  \boldsymbol{y} \left(t_{n}\right)\\
&+\frac{2  }{(3 \lambda_2 -\lambda_1) h} \boldsymbol{y} \left(t_{n-1}\right)
+\frac{- \lambda_1 -\lambda_2 }{2(3 \lambda_2 - \lambda_1 ) h}  \boldsymbol{y} \left(t_{n-2}\right)\\
&+\mathcal{O}(h^2).
\end{aligned}
\end{equation}
Using equation \eqref{EQdoty1} to discretize equation \eqref{IVP}, one obtains
\begin{equation}\label{EQyp1}
    \begin{aligned}
    \boldsymbol{y}_{n+1} = &\frac{3\lambda_2 - \lambda_1}{2\lambda_2}h\boldsymbol{f}(t_n, \boldsymbol{y}_n)+\frac{3(\lambda_1 + \lambda_2)}{4 \lambda_2}\boldsymbol{y}_n - \frac{\lambda_1}{\lambda_2} \boldsymbol{y}_{n-1} \\
    &+ \frac{\lambda_1 + \lambda_2}{4 \lambda_2} \boldsymbol{y}_{n-2}.
    \end{aligned}
\end{equation}
Letting $\lambda_2 = \lambda \lambda_1$ and reformulating equation \eqref{EQyp1}, we have a formal description of the ZeroSNet:
\begin{equation}\label{EQyp2b}
    \begin{aligned}
    \boldsymbol{y}_{n+1} = &\frac{3(1 + \lambda)}{4 \lambda}\boldsymbol{y}_n - \frac{1}{\lambda} \boldsymbol{y}_{n-1}  + \frac{1 + \lambda}{4 \lambda} \boldsymbol{y}_{n-2} \\ 
    &+ \frac{3\lambda - 1}{2\lambda}h\boldsymbol{f}(t_n, \boldsymbol{y}_n), ~~\lambda \neq 0.
    \end{aligned}
\end{equation}
\section{Proofs of Theorems}
We give proofs of all theorems in this work as follows. 
\subsection{Proof of Theorem 1}
\begin{proof}
    For a continuously differentiable function $\boldsymbol{y}(t)$, consistency conditions (7.7) and (7.8) in \cite{atkinson2011numerical} are respectively guaranteed for ZeroSNet \eqref{EQyp2} by
    \begin{equation}\label{ConsCondEQ1}
        \frac{3(1 + \lambda)}{4 \lambda}
        - \frac{1}{\lambda}
        + \frac{1 + \lambda}{4 \lambda} =1,
    \end{equation}
    and
    \begin{equation}\label{ConsCondEQ2}
         -(- \frac{1}{\lambda}  +  \frac{2(1 + \lambda)}{4 \lambda}) + \frac{3\lambda - 1}{2\lambda}
         =1.
    \end{equation}
\end{proof}
\subsection{Proof of Theorem 2}
\begin{proof}
    Consider the characteristic equation of \eqref{EQyp2}:
    \begin{equation}\label{EQce}
        \begin{aligned}
            r(\rho) &=\rho^{3}-\frac{3 \lambda+3}{4 \lambda} \rho^{2}+\frac{1}{\lambda} \rho-\frac{1+\lambda}{4 \lambda}.
        \end{aligned}
    \end{equation}
    Solving $r(\rho)=0$ gives a root which is always on the unit circle ($\rho_0=1$) and other two roots are given by
    \begin{equation}
        \rho_{1,2} (\lambda) = \frac{3-\lambda\pm \sqrt{(9+5\lambda)(1-3\lambda)} }{8 \lambda}.
    \end{equation}
    Let $\max\{|\rho_{1}|, |\rho_{2}| \}<1$, and then $\lambda \in (- \infty, -1) \cup (1/3, +\infty)$. According to consistency (provided by Theorem \ref{ThCons}) and Theorem 7.6 in \cite{atkinson2011numerical}, the ZeroSNet \eqref{EQyp2} is zero-stable in $\lambda \in (- \infty, -1) \cup (1/3, +\infty)$.
\end{proof}

\subsection{Proof of Theorem 3}

\begin{proof}
    For a zero-stable ZeroSNet \eqref{EQyp2} with characteristic equation \eqref{EQce}, if the discriminant $ \triangle \geq 0$, we have $\lambda \in [-9/5, -1)$. Consider squares of the two solutions \cite{irving2004integers}: $ \rho_1^2 = (3-\lambda + \sqrt{\triangle})^2/(64\lambda^2)$ and $ \rho_1^2 = (3-\lambda - \sqrt{\triangle})^2/(64\lambda^2)$. Combining the range of $\lambda$, one obtains that $\max \{ \rho_1^2, \rho_2^2\} = \rho_1^2 \in [1/9, 1)$ with its minimum taken at $\lambda=-9/5$.

    If $ \triangle < 0$, for a zero-stable ZeroSNet \eqref{EQyp2}, two complex-conjugate solutiuons satisfy $\rho_1^2 = \rho_2^2 = 1/4 + 1/(4 \lambda)$. Then, $\max \{ \rho_1^2, \rho_2^2\} \in (1/9, 1/4) \cup (1/4, 1)$. It can be derived that $\forall \lambda \in (-\infty, -9/5) \cup (1/3, \infty)$, $\rho(\lambda) > 1/3 = \rho(-9/5)$.

    Combining the above cases and substituting $\lambda=-9/5$ into equation \eqref{EQyp2} gives Theorem \ref{THEOW}.
\end{proof}

\end{document}